\documentclass[sigconf, dvipsnames,table,xcdraw,preprint]{acmart}

\usepackage{epsf}
\usepackage{graphicx}
\usepackage{hyperref}
\usepackage{xr-hyper}
\usepackage[ruled,vlined]{algorithm2e}
\usepackage{booktabs, makecell, tabularx}
\usepackage{amsmath}
\usepackage{mathtools}
\usepackage[labelsep=period]{caption}
\usepackage{soul} 
\usepackage{graphicx}
\usepackage{caption}
\usepackage{subcaption}
\usepackage{float}
\usepackage{multirow}

\makeatletter
\def\@copyrightspace{\relax}
\makeatother

 \setcopyright{none}
\renewcommand\footnotetextcopyrightpermission[1]{} 
\begin{document}

\title{Heed the Noise in Performance Evaluations in Neural Architecture Search}


\author{Arkadiy Dushatskiy}
\affiliation{%
  \institution{Centrum Wiskunde \& Informatica}
  \city{Amsterdam} 
  \country{the Netherlands} 
}
\email{arkadiy.dushatskiy@cwi.nl}

\author{Tanja Alderliesten}
\affiliation{%
  \institution{Leiden University Medical Center}
  \city{Leiden} 
  \country{the Netherlands} 
}
\email{t.alderliesten@lumc.nl}

\author{Peter A. N. Bosman}
\affiliation{%
  \institution{Centrum Wiskunde \& Informatica}
  \city{Amsterdam} 
  \country{the Netherlands} 
}
\affiliation{%
  \institution{Delft University of Technology}
  \city{Delft} 
  \country{The Netherlands} 
}
\email{peter.bosman@cwi.nl}

\renewcommand{\shortauthors}{Dushatskiy, et al.}

\begin{abstract}
 Neural Architecture Search (NAS) has recently become a topic of great interest. However, there is a potentially impactful issue within NAS that remains largely unrecognized: noise. Due to stochastic factors in neural network initialization, training, and the chosen train/validation dataset split, the performance evaluation of a neural network architecture, which is often based on a single learning run, is also stochastic. This may have a particularly large impact if a dataset is small. We therefore propose to reduce this noise by evaluating architectures based on average performance over multiple network training runs using different random seeds and cross-validation. We perform experiments for a combinatorial optimization formulation of NAS in which we vary noise reduction levels. We use the same computational budget for each noise level in terms of network training runs, i.e., we allow less architecture evaluations when averaging over more training runs. Multiple search algorithms are considered, including evolutionary algorithms which generally perform well for NAS. We use two publicly available datasets from the medical image segmentation domain where datasets are often limited and variability among samples is often high. Our results show that reducing noise in architecture evaluations enables finding better architectures by all considered search algorithms.
\end{abstract}



\maketitle
\pagestyle{plain}
\section{Introduction}

\subsection{Neural architecture search}
Neural Architecture Search (NAS), i.e., the automated design of Neural Networks (NNs) architectures tailored for a specific task, has become a topic of great interest recently. The main reason for that is the growing number of ideas in NN design, many of which demonstrate great performance. However, with this growth, it is becoming more difficult to guess without running experiments which network would be the best for a given task. This makes automated network design a natural research topic positioned in between deep learning and optimization algorithms. 

Optimization algorithms used for NAS include Evolutionary Algorithms (EAs) \cite{ci2021evolving, so2019evolved, yu2020c2fnas}, Bayesian optimization algorithms using performance predictors \cite{shi2020bridging, white2020local}, gradient descent-based methods \cite{liu2019auto, yan2020ms}, reinforcement learning algorithms \cite{gao2019graphnas}, and Local Search (LS) \cite{den2021local, white2020local}. Gradient descent-based methods use a so-called supernetwork, of which the structure is optimized using a gradient descent optimizer simultaneously with the network weights. However, it was shown in \cite{yang2019evaluation} that the performance of such methods is often suboptimal, in some cases not better than the most simple search approach - random search. 

NAS can be extremely computationally expensive if the search relies on training numerous networks. To reduce the computational costs, part of the network trainings can be replaced by a computationally cheaper performance estimation made by performance predictors (also called surrogate models). However, developing a powerful performance predictor for NAS might be challenging and search space specific \cite{white2021powerful}. 

\subsection{Medical image segmentation}
Segmentation is one of the major tasks in computer vision. Given an image, the task is to automatically perform a specific pixel-wise classification that outlines certain things in the image. Medical Image Segmentation (MIS) is a special case with input images being medical scans, such as Magnetic Resonance Imaging (MRI) or Computed Tomography (CT) acquired of a region of interest determined by a medical expert. Examples of common MIS tasks are organ, tumour, and vessel segmentation. Designing a precise and fast MIS algorithm can be immensely beneficial for healthcare as potentially it can not only reduce the workload of physicians (for manual scan segmentation), but also make some medical procedures faster which may be beneficial for patients as well. In some cases, segmentation performed by a deep NN can demonstrate human-level performance \cite{nikolov2018deep}. Most of the proposed architectures are adaptations of Unet \cite{ronneberger2015u}. The main idea of Unet is an encoder-decoder structure which, firstly, extracts features from the image, and then, translates them into a segmentation. Despite good performance of Unet in general, due to variability in both image quality and expert input segmentations, finding the best network architecture for a specific task is still challenging, and therefore, NAS for MIS has high practical value.

\subsection{Neural architecture search for medical image segmentation}
Research on NAS for MIS has so far been less elaborate than on NAS for classification. However, several works have shown that automatically found networks perform better than the state-of-the-art manually designed ones \cite{yan2020ms, yu2020c2fnas, weng2019unet, isensee2018nnu}. In general, the search can be performed on two levels: 1) configuration and combinations (also called \emph{a cell}) of atomic operations in a network (e.g., number and kernel size convolutions, or activation functions); 2) network topology, i.e., defining the connections between cells, and input /output tensor dimensionalities for them. In \cite{weng2019unet} NAS-Unet was introduced, the main idea of which is to search for the configuration of cells in the fixed Unet structure. The best found network demonstrated better performance than a standard Unet. In \cite{yu2020c2fnas} it was shown that a bilevel search of a Unet-like network topology at the first level and the structure of cells at second level, can find even better performing networks than NAS-Unet. Simultaneous search of cell structure and network topology was performed in \cite{yan2020ms}, also showing better performance than NAS-Unet. This suggests that network topology and the configuration of cells are connected. In \cite{isensee2018nnu} state-of-the-art performance was achieved on 10 MIS tasks from the Medical Segmentation Decathlon challenge \cite{antonelli2021medical} by a newly proposed method called nnUnet. However, it can be considered to be a semi-NAS method as, while the networks are automatically configured for each task, the number of options is very limited and the network construction is rule-based, determined by the dataset properties such as the resolution of the scans.

\subsection{Potentially impactful issue: noise}
NAS, just as any search task, needs a definition of a function that maps a solution (i.e., a candidate network architecture) to a performance score. It is common practice in NAS to evaluate a network (assign a performance score) using a fixed validation set. Moreover, in most cases, the evaluation is deterministic, meaning that only one random seed for the initialization of network weights and stochastic training components (i.e., applied augmentations and sampling of batches) is used. 

\begin{figure}[h!]
    \centering
    \includegraphics[width=0.45\textwidth]{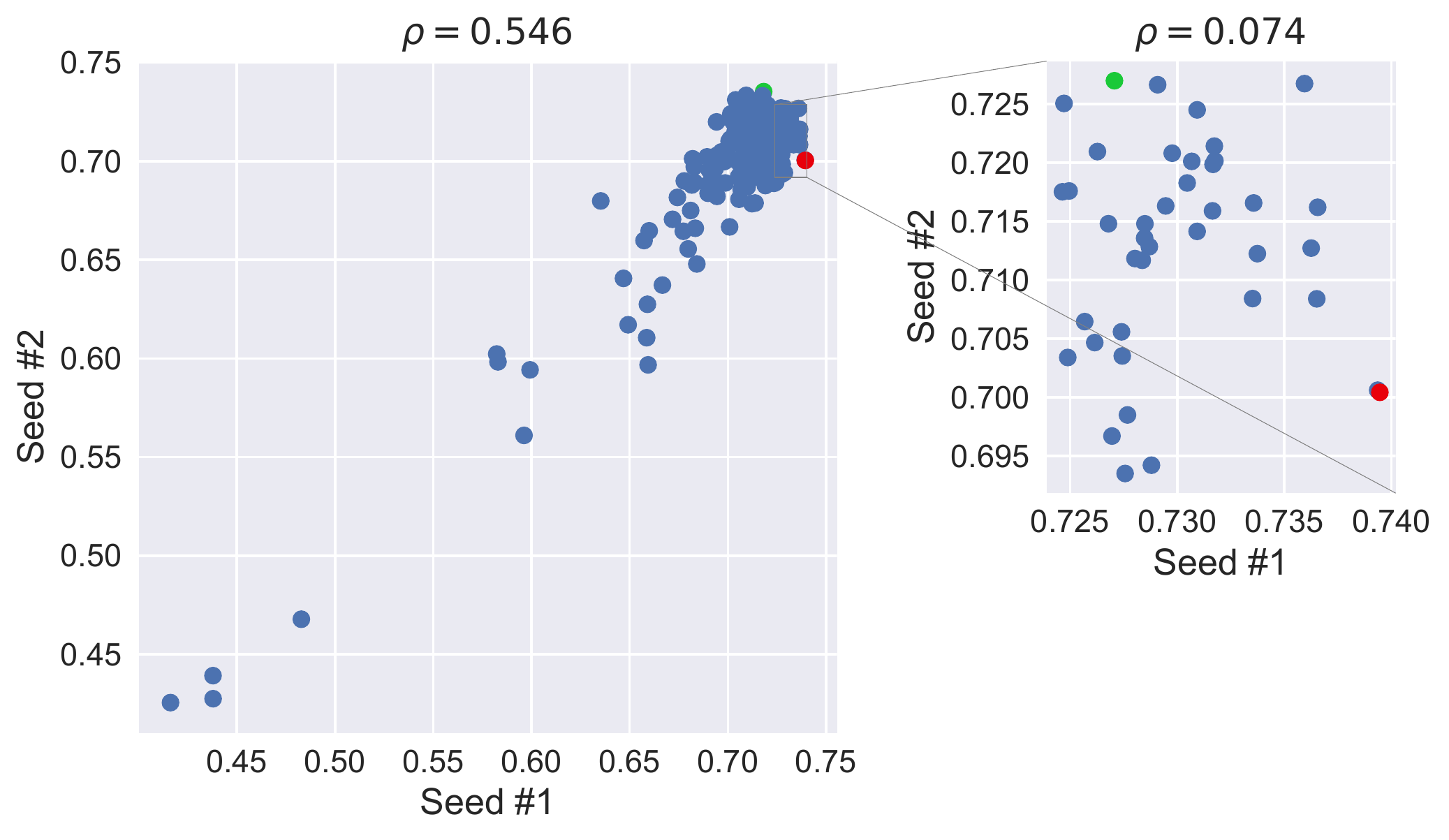}
    \includegraphics[width=0.45\textwidth]{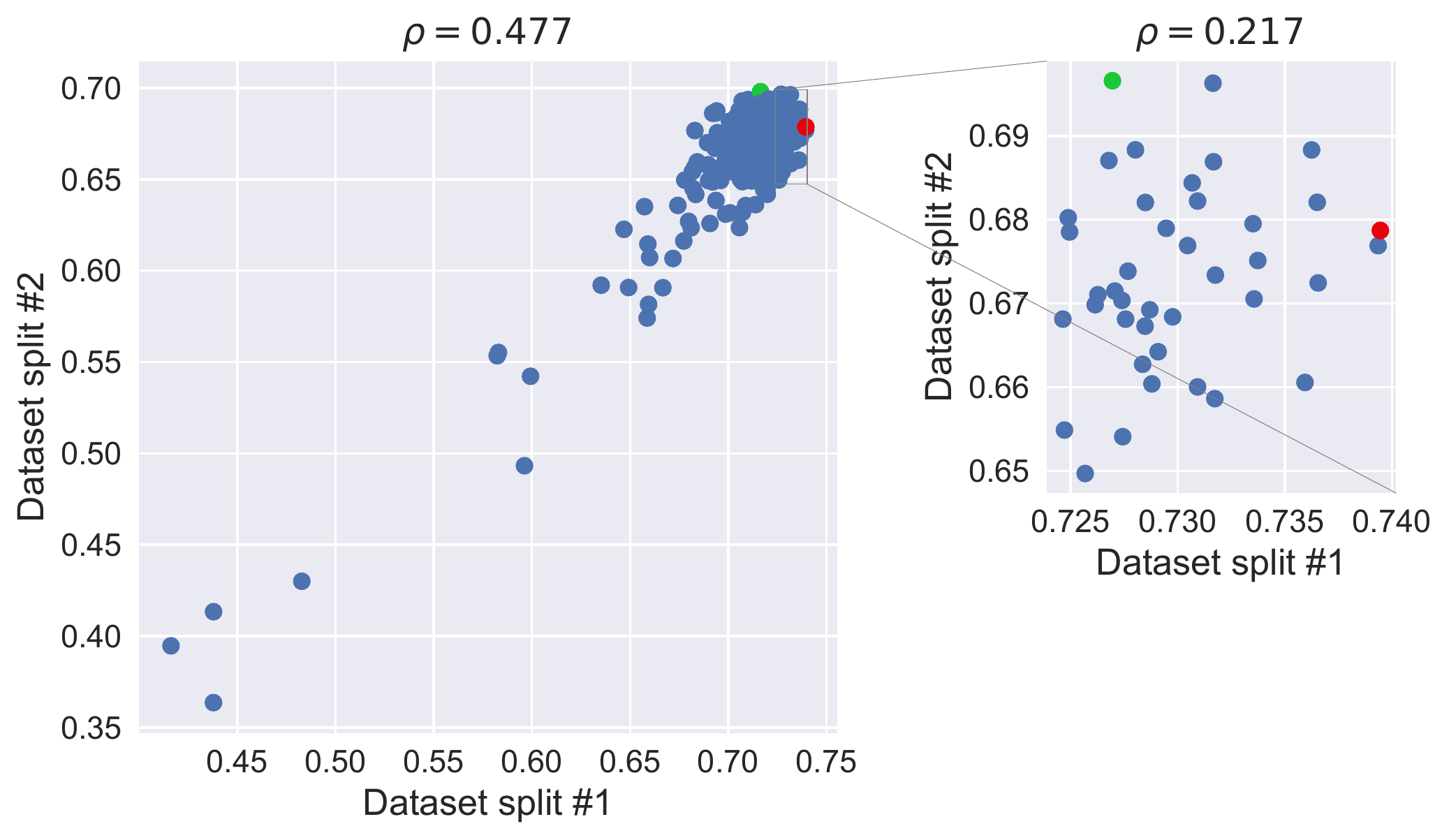}
    \caption{Differences in performance of NNs trained with different random seeds (upper plot) and different train/validation data splits (lower plot). Each dot represents one architecture. Zoomed-in plots correspond to the top 20\% of the networks (that will likely be discovered in the later stages of the search process). Spearman rank correlation is denoted by $\rho$. The dataset is the Prostate segmentation dataset from the Medical Segmentation Decathlon \cite{antonelli2021medical}. The performance metric displayed in both axes is the Dice coefficient (see definition in Section~\ref{sec:metric}). Red and green dots denote the best architectures according to each seed (top) and each train/validation data split (bottom). Networks in the plots are collected during a run of the SAGOMEA search algorithm (see Section ~\ref{sec:searchalgorithms}) with a budget of 200 function evaluations.}
    \label{fig:example_seeds}
\end{figure}

\begin{figure}[h]
    \centering
    \includegraphics[width=0.4\textwidth]{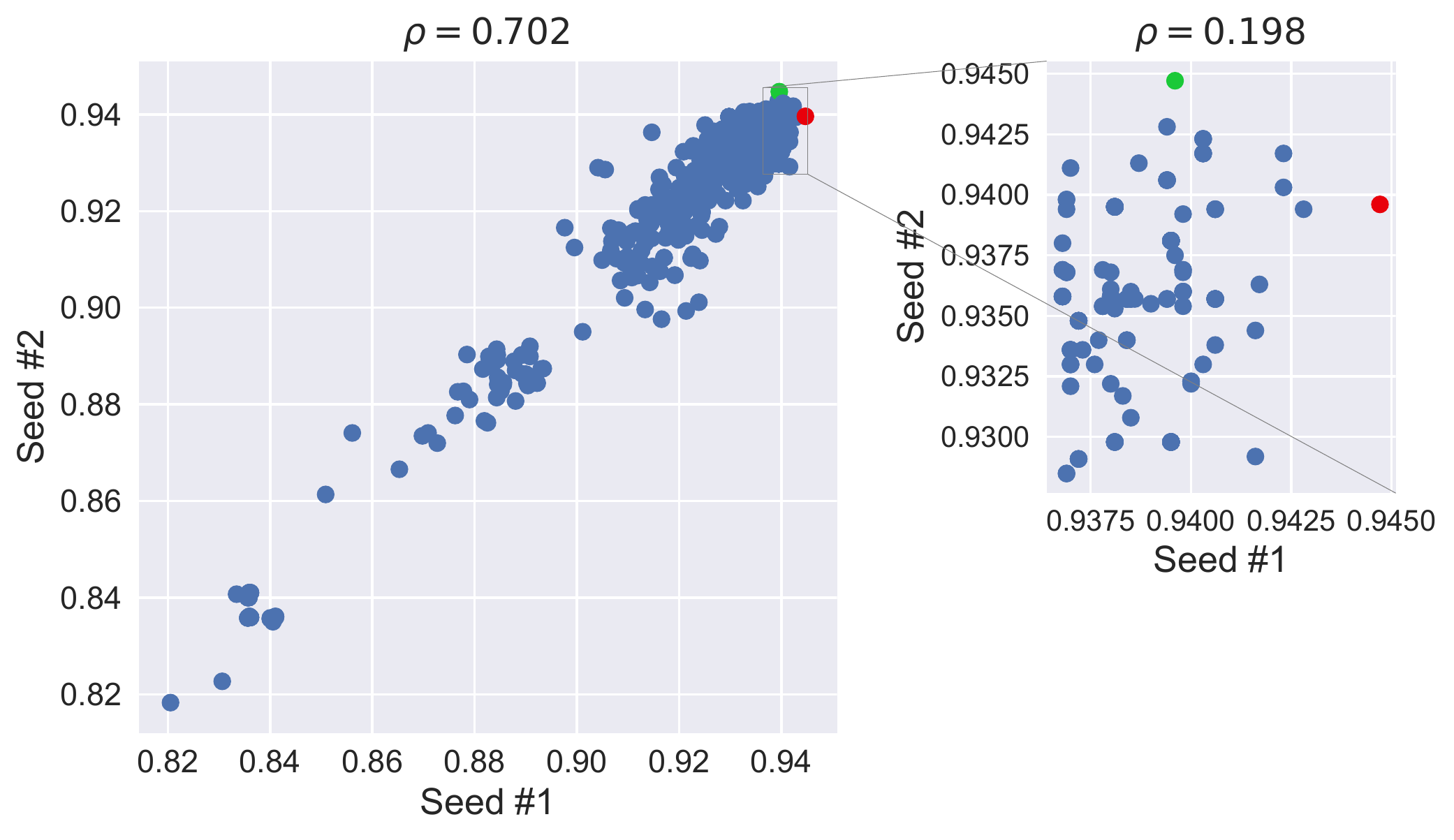}
    \caption{ Differences in performance of NNs trained with different random seeds from the NAS-101 tabular benchmark (CIFAR-10 dataset). Each dot represents one architecture. The zoomed-in plot corresponds to the top 20\% of the networks (that will likely be discovered in the later stages of the search process). Spearman rank correlation is denoted by $\rho$. The performance metric displayed in both axes is validation accuracy. Red and green dots denote the best architectures according to each seed. Networks in the plot are collected during a run of the Local Search algorithm (see Section ~\ref{sec:searchalgorithms}) with a budget of 1000 function evaluations.}
    \label{fig:example_seeds_cifar}
\end{figure}

In our own experiments with NAS, we have noticed that, NN network performance can depend a lot on the chosen random seed and train/validation dataset split, especially (in a relative sense) when considering well-performing architectures. A demonstration of this for the case of MIS is given in Figure~\ref{fig:example_seeds}. The Spearman rank correlation between networks trained with two different random seeds is poor ($<0.1$) when calculated for the top 20\% of the networks. A similar result applies for networks trained with two different train/validation data splits. Moreover, the best network according to one random seed or one train/validation data split, does not correspond to the best network when another seed or train/validation data split is used. This means that using a standard evaluation procedure (one seed and one train/validation data split) might lead to finding suboptimal networks, of which the true performance (e.g., on an independent test set or obtained by cross-validation with several random seeds if a test set is not available) is not as good as expected. Similar results are shown for the case of NAS for classification using the NAS-101 benchmark \cite{nas101} in Figure \ref{fig:example_seeds_cifar}.

In this paper, we study whether increasing the network performance evaluation reliability by using cross-validation and multiple random seeds leads to better quality of networks found by NAS. We use different performance evaluation setups as fitness functions for various search algorithms to understand whether more computationally expensive network evaluations lead to better generalization. To the best of our knowledge, this is the first time such a study is conducted. In contrast to just measuring network performance using different training random seeds (as, for instance, in NAS-101 \cite{nas101}, NAS-201 \cite{nas201}), we study how more reliable performance evaluation setups affect the found network quality when measured in an \emph{independent} performance evaluation. We focus on NAS for MIS as medical image datasets used for training segmentation models are often small (tens of scans) and the generalization problem, i.e., the problem of results transferability obtained for a particular training random seed, or validation subset, might have a more substantial impact on NAS.

\section{NAS method}
To use a NAS method, a search algorithm and a fitness function need to be defined. The choice of search algorithm depends on the chosen NAS task formulation (e.g., a combinatorial optimization paradigm). The definition of the fitness function depends on the chosen performance evaluation strategy and the selected segmentation quality metric.

\subsection{Search algorithms} \label{sec:searchalgorithms}
The first considered search algorithm is Local Search (LS). It is a simple search approach, but it was shown to perform better in various NAS cases than random search and in some cases even be on par with advanced EAs \cite{den2021local, white2020local}. LS works by iterating over variables in a random order and greedily choosing the best option for each variable. Evolutionary algorithms are powerful general-purpose search algorithms and have also successfully been applied to NAS, see, e.g., \cite{real2019regularized, so2019evolved}. Therefore, the next algorithm we consider is the parameterless version of the state-of-the-art EA Gene-pool Optimal Mixing Evolutionary Algorithm (P3-GOMEA) \cite{dushatskiy2021novel, dushatskiy2021parameterless} (further referred to as GOMEA). P3-GOMEA is a model-based EA that attempts to detect and exploit linkage information during optimization. It was shown to perform better than other EAs on a range of problems \cite{dushatskiy2021novel}, and, importantly, it does not have a population size hyperparameter which needs to be tuned in many other EAs. Further, we consider surrogate-assisted GOMEA (SAGOMEA): a modification of GOMEA which uses a surrogate model for cheap fitness estimation. SAGOMEA was designed specifically for discrete optimization problems with computationally expensive fitness functions and has been shown to be efficient \cite{dushatskiy2021novel}. Finally, we use a Bayesian optimization algorithm called Tree Parzen Estimator (TPE). Specifically, we use its implementation in the Hyperopt optimization package \cite{hyperopt}. Similarly to SAGOMEA, TPE was designed for problems with expensive fitness functions.

\subsection{Segmentation quality metrics} \label{sec:metric}
To use the above-described search algorithms, a fitness function needs to be defined. The Dice score is a commonly used metric for medical image segmentation quality evaluation. There exist other metrics to evaluate segmentation quality (Surface Dice coefficient, Hausdorff distance), but in this work we focus on the most commonly used in literature Dice coefficient. Moreover, its calculation is computationally cheap and independent from from the voxel spacing of scans. For the Dice coefficient calculation, a reference segmentation $R$ and the predicted segmentation $P$ need to be binary segmentation maps of dimensionality $C \times X \times Y \times Z$ (where $C$ is the number of segmentation classes, and $X\times Y \times Z$ is the scan size), i.e., the value one in position $(c,x,y,z)$ means that the voxel with coordinates $(x,y,z)$ belongs to class $c$. We use the average Dice of multiple classes which is defined by the formula $\frac{1}{C-1} \sum_{1}^{C} \frac{2|R_c \cap P_c|}{|R_c|+|P_c|}$. Note that the zero class which is usually the image background is not included in the Dice calculation.

\subsection{NAS task formulation}
We formulate the NAS task as a combinatorial optimization problem (maximization) with the search space consisting of possible network architectures encoded with discrete variables and the fitness function being a network segmentation performance (in our case: the Dice coefficient). Ultimately, the goal of NAS is to find an architecture which performs well on unseen (during search) data. Therefore, it is a common practice in NAS, to divide the dataset into train, validation, and test parts. In each fitness function evaluation of NAS, a network is trained (from scratch) on the train subset, and its performance is evaluated on the validation subset. After obtaining the best performing network according to the validation set, i.e., at the end of NAS, its performance is verified on the test subset. This way, it is checked whether overfitting to the specific validation subset took place. However, in case of small medical datasets, a performance on a single test subset might be not sufficiently indicative. Therefore, we do not use a separate test set. Instead, for the final evaluation procedure (to obtain the true performance measure), we retrain the networks from scratch using an averaged cross-validation performance on previously (during search) unseen cross-validation data splits and random seeds different from the ones used for training during optimization.

\subsection{Network performance evaluation} \label{sec:performanceestimation}
\begin{figure}[ht]
    \centering
    \includegraphics[width=0.35\textwidth]{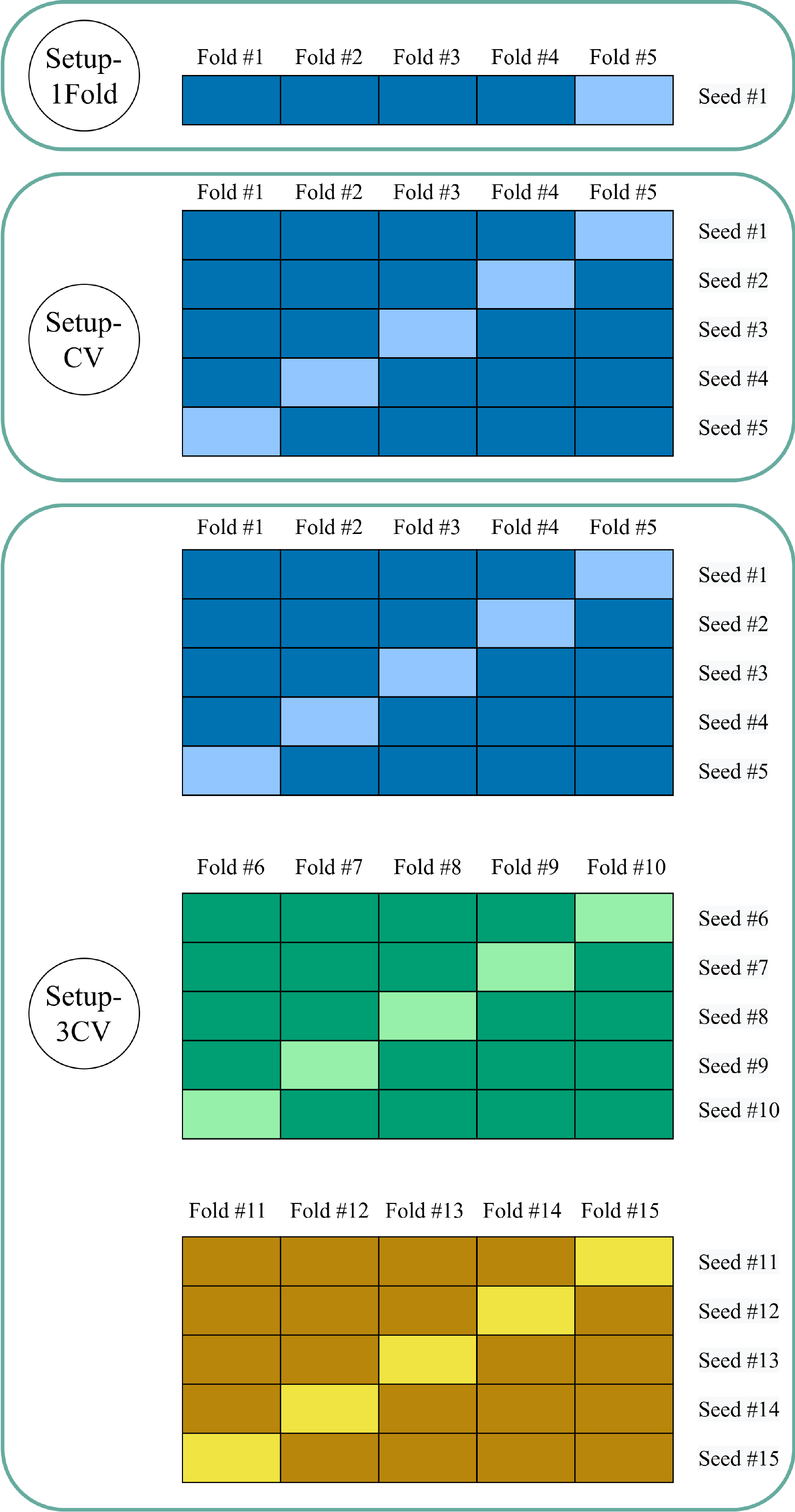}
    \caption{Different setups for evaluating architecture performance. Each row denotes one training. In each row, a lighter rectangle denotes the used validation subset, while darker ones constitute a training subset. The second and the third data splits used in Setup-3CV (with fold numbers 6-10, 11-15) denote different cross-validation data splits (this is also shown with different colors) than the one with the fold numbers 1-5. In total, Setup-1Fold requires one network training, Setup-CV 5, and Setup-3CV 15.}
    \label{fig:types}
\end{figure}

The most basic way to evaluate network performance (i.e., map a solution to a fitness score), is to calculate the Dice score on a single validation set using one random seed for network initialization and training. However, due to the differences between medical scans, performance of networks may vary a lot depending on the validation dataset. Moreover, due to the stochastic nature of network training and initialization, performance may depend on the random seed used for weights initialization and training. This may affect NAS. To mitigate this problem, we investigate different possible evaluation procedures. The first considered procedure is a basic one that is mostly used in NAS literature: only one random seed and a fixed validation set (\textbf{Setup-1Fold}). The second proposed evaluation procedure uses 5-fold cross-validation (CV) instead of a single validation set (\textbf{Setup-CV}). In order to address both of the above-mentioned reasons for performance variance, different random seeds are used for network initialization and training in each of the cross-validation folds. The most computationally expensive considered evaluation procedure repeats 5-fold cross-validation using three different data partitionings into folds (\textbf{Setup-3CV}). These three evaluation procedures are shown schematically in Figure~\ref{fig:types}.

\section{Search space}
In choosing the search space for our experiments, we followed three main design principles: 1) The search space should contain networks that perform reasonably well, preferably better than a Unet. 2) The search space should be large and contain diverse networks to make the search a non-trivial task. 3) The search space should not contain networks which are prohibitively computationally expensive to train. We decided to adopt the search space used in \cite{yu2020c2fnas} while making it more versatile to meet all these criteria. Such a search space allows NAS formulation as a discrete combinatorial problem, is reasonably large, and in \cite{yu2020c2fnas} promising results are shown. Instead of a bi-level search, we search simultaneously for the architecture topology and cells. Secondly, we allow different cells instead of repeating the same structure in all positions. Finally, to enlarge the search space and avoid the situation that the majority of networks are considered infeasible, we lift the architectural restrictions which were applied in \cite{weng2019unet}: in contrast to a fixed encoder-decoder structure, we allow all possible architectures from the topology space described below.
\subsection{Topology search subspace}  \label{sec:topsearchspace}
\begin{figure*}[ht]
    \centering
    \includegraphics[width=0.97\textwidth]{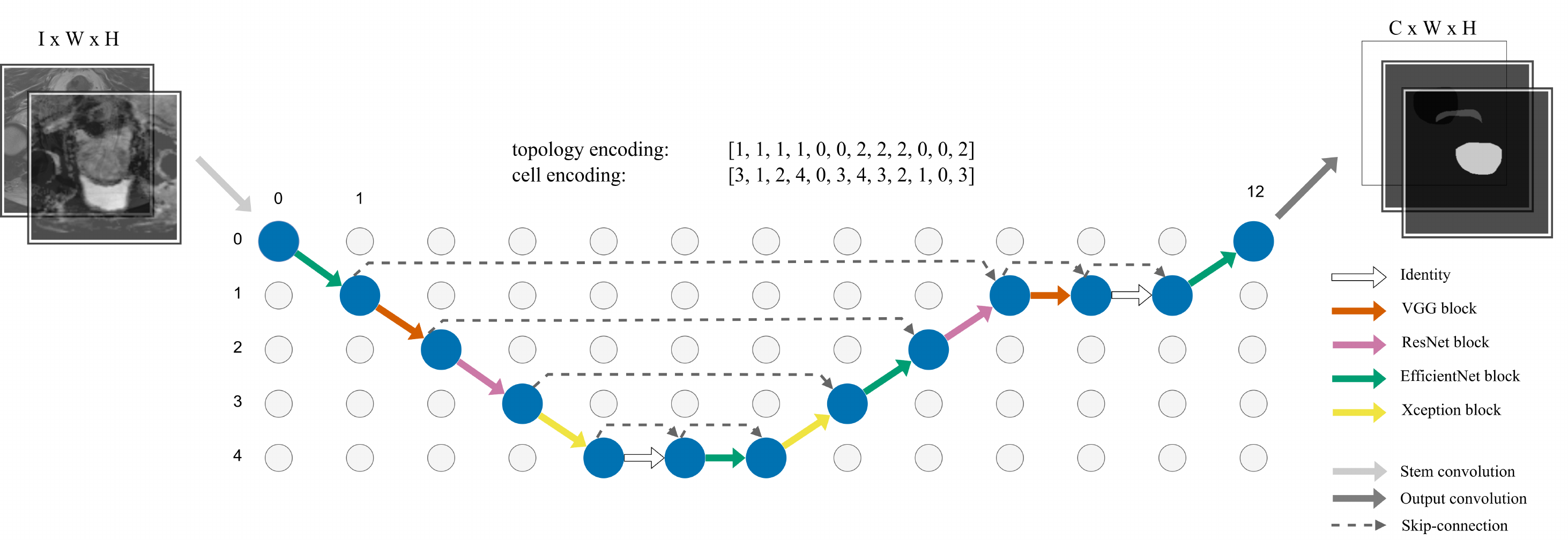}
    \caption{The scheme of network architectures in the proposed search space and an example Unet-like architecture. Each circle represents feature maps (a three-dimensional tensor), the blue ones are used in the shown architecture. Feature maps at depth $j$ (from 0 to 4) have $2^j$ smaller spatial resolution and $2^j$ more channels than the feature maps after applying stem convolution (in the left upper corner). Each architecture is defined as a sequence of $N=12$ cells (represented by arrows). Each cell is defined by two discrete variables: its main block specification (five different options), and its type: normal (norizontal arrows), downsampling (downward arrows), or upsampling (upward arrows). Skip-connections are added using fixed rules (see Section~\ref{sec:searchspace3}). A stem convolutional block translates an input image to feature maps with 32 channels. An output convolutional block translates corresponding feature maps to segmentation masks.}
    \label{fig:searchspace}
\end{figure*}

The topology part of the search space determines the general structure of the network. It is shown in Figure~\ref{fig:searchspace}. The network topology is defined by a sequence of $N$ cells, which can be on different levels, i.e., $S=(l_0, l_1, \dots, l_{N-1})$. At level $i$, feature maps have the dimensionality $\left (D*2^i, \frac{W'}{2^i}, \frac{H'}{2^i} \right)$, where $D$ is the number of channels of the network input (after the stem convolution is applied, see Section~\ref{sec:searchspace3}), and $(W',H')$ is its spatial dimensionality. We allow only one level change between the consecutive cells: i.e., $|l_i-l_{i+1}| \le 1$. Therefore, to encode the network topology $S$, it is sufficient to specify whether to increase the level (from $l_i$ to $l_i+1$) for the next cell (\emph{downsampling}), decrease (\emph{upsampling}), or keep it at the previous level (\emph{normal}). The topology search space encoding is therefore a vector of size $N$: $\{0,1,2\}^N$, where $0,1, \textrm{and}~2$ encodes normal, downsampling, and upsampling cells respectively.
Note that some vectors from such search space represent infeasible architecture topologies as 1) it is not possible to apply an upsampling operation from the first level, and 2) a downsampling operation cannot be applied from the maximum allowed level. The most straightforward way to fix a vector representing an infeasible architecture is to change infeasible downsampling or/and upsampling operations to normal ones.
\begin{figure}[h]
    \centering
    \includegraphics[width=0.47\textwidth]{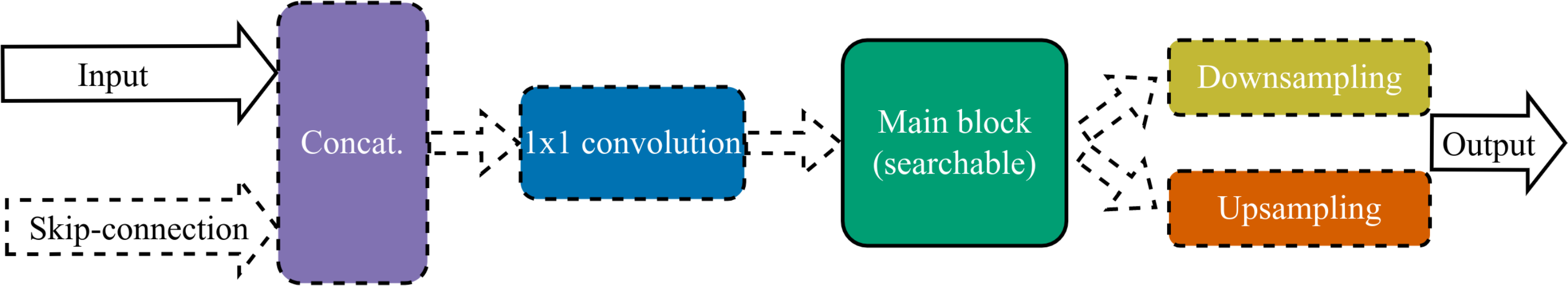}
    \caption{The network cell structure. The dotted parts are optional. Downsampling and upsampling operations depend on the network topology, while a skip-connection is always applied when possible (see Section~\ref{sec:cellssearchspace}). 
    When a skip-connection is applied, the corresponding feature maps are firstly concatenated (Concat. block) with the input feature maps and then passed through a $1\times1$ convolution to keep the spatial dimensions equal to the input dimensions.
    }
    \label{fig:cell}
    \vspace{-0.3cm}
\end{figure}
\subsection{Cells search subspace} \label{sec:cellssearchspace}
The structure of a cell is shown in Figure~\ref{fig:cell}. Each cell consists of its main block, optional downsampling or upsampling operations, and an optional skip-connection from one of the previous cells. As the necessity for downsampling or upsampling is encoded in the topology subspace, and the skip-connections presence is fixed (see Section~\ref{sec:searchspace3}), only block types need to be additionally encoded.

For each cell block, we consider five different options, four of which are non-trivial and one is an identity operation, the purpose of which is to act as a placeholder and allow more light-weight networks. The first type of block is a VGG block \cite{simonyan2014very}, which is also used in a standard Unet \cite{ronneberger2015u}. In addition to this type of block, we consider three blocks that are used in classification networks that exhibit top performance: a block with a residual connection \cite{he2016deep, wightman2021resnet} (ResNet block; a block that uses a concept of a depth-wise separable convolution \cite{chollet2017xception} (Xception block); and a block used in one of the state-of-the-art classification networks, namely, Efficient-net \cite{tan2019efficientnet}. We believe that the selected options can, firstly, contribute to a good performance in segmentation tasks, and, secondly, make the architectures in the search space diverse as different blocks can be used in different parts of the network.

After performing block operations, downsampling cells perform convolution with stride 3 and stride 2. In the upsampling cells, the upsampling operation is transpose convolution with kernel size 3.

Similar to \cite{yu2020c2fnas}, skip-connections are added to the input of a cell in two cases: 1) The cell follows an upsampling cell. Then, a skip-connection is added from the previous cell on the same level. 2) There is a cell before the previous one on the current level of input. Then, a skip-connection is added from that cell. 

\subsection{Search space details} \label{sec:searchspace3}
Just as done in related literature \cite{yu2020c2fnas, liu2019auto}, we set the maximum network depth to $N=12$. Note, however, that in contrast to \cite{yu2020c2fnas}, we allow the number of effective cells to be lower due to possible identity blocks. The maximum cell level is $5$, this value was also used in \cite{yu2020c2fnas}. 
In total, our search space encoding consists of $N=12$ discrete variables which encode the network topology, and 12 variables which encode the corresponding block type for each cell in the network.
The topology related variables have a cardinality of 3, while the cell type variables have a cardinality of 5.

\section{Experiments}

First, we compare different search algorithms in terms of performance on the given optimization problems. Secondly, we compare how the quality of the found architectures after independent re-training and re-evaluation depends on the used performance evaluation approach. Statistical tests (Wilcoxon test with Bonferroni correction, $\alpha=0.05$) are conducted to verify the results. Then, we analyze the reasons for the observed differences. Finally, we compare performance of the best found networks to commonly used handcrafted architectures.

\subsection{Experimental setup} \label{sec:exprs}

\subsubsection{Performance evaluation setups} We consider three setups for the evaluation of neural network performance during the search (i.e., the fitness functions) as described in Section~\ref{sec:performanceestimation}: 1) Training on one seed and using one dataset split (Setup-1Fold); 2) Using 5-fold cross-validation (Setup-CV); 3) Using 5-fold cross-validation on three different partitionings of the folds (Setup-3CV). For each of the setups and each of the algorithms (LS, GOMEA, TPE, SAGOMEA) we perform five runs per dataset. These runs differ in both the random seed used by the search algorithm, and the seeds used for the networks performance evaluation. 

\subsubsection{Handling infeasible solutions} For all search algorithms, the same strategy of handling solutions that encode infeasible architectures is applied: before evaluating fitness, an architecture is checked for feasibility, and, if necessary, repaired as described in Section~\ref{sec:topsearchspace}. Note that the algorithms receive a fitness score for infeasible solutions, i.e., the repaired genotype is only used for evaluation; it does not replace the original infeasible genotype. Such choice was made in order to fairly compare different search algorithms without modifying them for infeasible solutions handling.

\subsubsection{Independent performance measure} The final evaluation metric (also further referred to as \emph{architecture quality}) is the performance of the architecture obtained after retraining it from scratch with three different 5-fold cross-validations (basically, as in the Setup-3CV). Importantly, both random seeds and cross-validation splits in this independent training and evaluation procedure do not overlap with the seeds and splits used during the search phase. While averaging only three cross-validations preserves some amount of noise in the score, we did not observe a substantial change in results if more than three cross-validations are used, and, therefore, we stick to the computationally cheaper procedure of using three different cross-validations.

\subsubsection{Computational budget} \label{subsubsec:compbudget} For all setups we allocate an equal computational budget in terms of network trainings. We study performance under four different computational budgets: \emph{T,~2T,~4T,~8T}. Due to computational time constraints, the largest considered budget of 8T comprises 3000 network trainings. Smaller budgets of T, 2T, 4T comprise 375, 750, 1500 network trainings correspondingly. With the Setup-1Fold, one fitness evaluation entails one network training. Thus, given, for instance, budget T, search algorithms perform T fitness evaluations in each optimization run. As the Setup-CV entails 5 network trainings in each fitness evaluation, search algorithms perform T/5 fitness evaluations given budget T (which equals to the same number of network trainings, namely, T). With the Setup-3CV, which is 15 times more computationally expensive than the first one, search algorithms can perform only T/15 fitness evaluations under budget T. While running search algorithms, we count only actual network trainings by storing all trained networks (after repair) along with their measured performance in an archive.

\subsection{Datasets}
In this work we perform experiments on two publicly available medical image segmentation datasets. The first segmentation task is prostate segmentation taken from the Medical Segmentation Decathlon challenge \cite{antonelli2021medical}. This dataset consists of 30 MRI scans (two scans with ids 18 and 32 were removed from the original dataset due to suspected label inconsistencies compared to the other scans) comprising two modalities: T2-weighted and ADC. The segmentation classes are background, prostate peripheral zone, and prostate transition zone. The second dataset is from the Automated Cardiac Diagnosis Challenge (ACDC) \cite{bernard2018deep}. It contains MRI scans (in one modality) of 100 patients and the task is multi-class segmentation (left ventricular endocardium, myocardium and right ventricular endocardium). We keep only one scan per patient (from the diastole phase) to make the validation process easier, i.e., the total considered number of scans is 100. These datasets were chosen due to their diversity (were collected for different segmentation challenges, are focused on different organs) and relatively low scan resolution which allows for faster experiments. 

\subsection{Preprocessing and training}
We adopt the preprocessing resampling and voxel value normalization) and training setups used in the nnUnet framework \cite{isensee2018nnu} as it demonstrates state-of-the-art performance on various datasets. The loss function for training is a sum of soft Dice loss and cross-entropy. The optimizer is Stochastic Gradient Descent (SGD) with Nesterov momentum and weight decay. Momentum and weight decay values are 0.99 and $3*10^{-5}$ respectively. The initial learning rate is $0.01$ and a polynomial decay learning schedule is used. To avoid overfitting, data augmentations are used with magnitude and probability values adopted from the nnUnet. For computational efficiency reasons, in our main experiments for training we use patches of size $128\times128$ pixels which are randomly sampled from the original images. 
The training is performed for 40 epochs. This value was chosen as a trade-off between training time and network performance. 
In order to minimize noise coming from one of the performance factors that we do not focus on here, i.e., different performance scores after slightly different number of training epochs, we average the segmentation prediction (which is then evaluated) over the last five epochs.

\subsection{Implementation details}
Neural network training and evaluation is implemented in Python using Pytorch. LS, GOMEA, and SAGOMEA are implemented in C++. The source code is provided at \footnote{\href{https://github.com/ArkadiyD/Noise\_in\_NAS}{https://github.com/ArkadiyD/Noise\_in\_NAS}}.
Experiments are conducted on a system with Nvidia A100 GPUs. One full network training (40 epochs) and evaluation takes $\approx2$ minutes. Therefore, one search run with the largest budget of 8T takes approximately 100 GPU-hours.

\section{Results}

\subsection{Search performance}
First, we study how different search algorithms perform on given optimization tasks. These results are shown in Figure~\ref{fig:opt_results} (Prostate dataset, budget 4T), and in Supplementary, Figure~\ref{fig:budgets12} (budgets T and 2T). All results are also provided in tabular form in Supplementary, Table~\ref{tab:opt_results1}. LS and GOMEA are approximately equal in optimization quality on all three setups. TPE finds solutions with better average fitness values than LS and GOMEA on all three setups. The difference is more substantial with Setup-1Fold. SAGOMEA outperforms TPE on Setup-1Fold and Setup-CV, however, it slightly underperforms on Setup-3CV. Statistical significance testing results are provided in the Supplementary,~Table~\ref{tab:pvalues2}. With a budget of 4T, TPE performs better than LS and GOMEA on Setup-1Fold and Setup-3CV with statistical significance. SAGOMEA performs better than LS, GOMEA, and TPE on Setup-1Fold and Setup-CV with statistical significance. These results demonstrate that algorithms that use surrogate models and were designed for expensive optimization problems have great potential in NAS. In further experiments (the second dataset and larger budget) we use only SAGOMEA and TPE algorithms as they demonstrate better ability to solve the considered NAS optimization task.
\begin{figure}[h]
    \centering
    \includegraphics[width=0.47\textwidth]{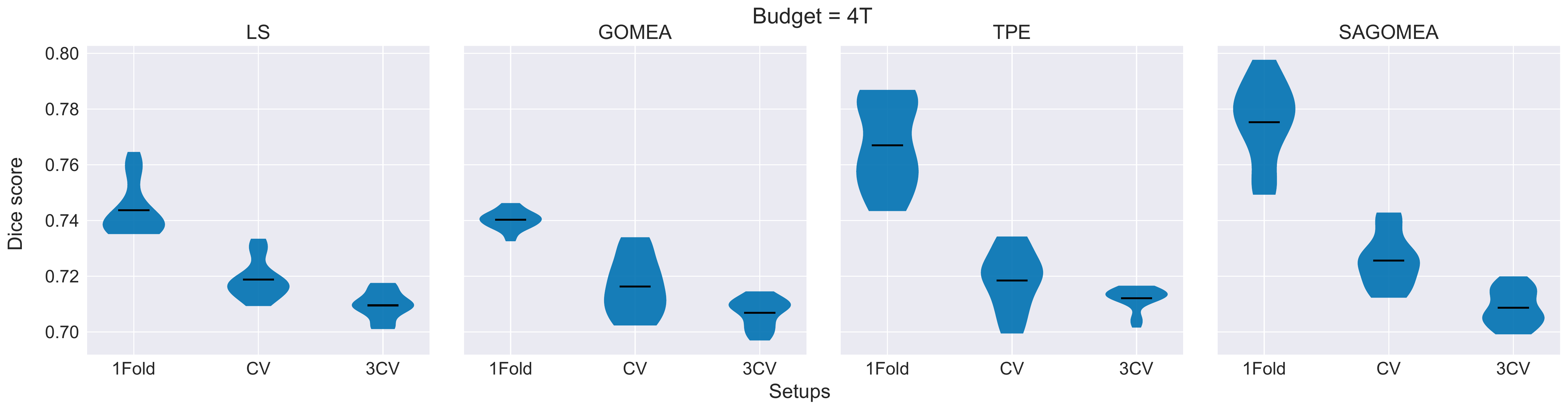}
     \caption{Comparing optimization performance of search algorithms in different setups with a computational budget of 4T (as described in Section~\ref{subsubsec:compbudget}). The image shows distributions (over five runs, five best architectures from each run) of search results. Note that these results take into account only fitness function values, not the final architectures quality. 
     }
    \label{fig:opt_results}
    
\end{figure}

\subsection{Quality of found networks}
Secondly, we study the differences in quality of the found architectures by different search algorithms with different performance evaluation setups and computational budgets. Quality of the networks is calculated independently from the search runs as described in Section~\ref{sec:exprs}. The results are shown in Figure~\ref{fig:search_results}; Supplementary, Figure~\ref{fig:main_results2} and in tabular form in Supplementary, Table~\ref{tab:quality_results}. These results suggest that with enough computational budget for a search run, it is better (on average) to use more reliable, yet computationally more expensive performance evaluation setups as the fitness function. We see that as the budget increases, in most cases Setup-CV starts to find better architectures than Setup-1Fold. We hypothesize that with an even larger budget the most computationally expensive Setup-3CV might level to even results than Setups-1Fold and Setup-CV as it demonstrates steady improvement with the budget increase. On the contrary, the performance of Setup-1Fold seems to improve slower or, in some cases (for instance, with TPE, on both datasets), even decline as the budget increases. Such behaviour indicates overfitting to a specific train/validation split and random seed. Similar trends are observed for all considered search algorithms which suggests that our findings do not depend on a specific search algorithm. Note that even with the budget of 8T, search algorithms with Setup-3CV are allowed to do only 200 function evaluations which is not a big number even for expensive optimization algorithms. From a practical perspective, the Setup-CV seems to be a good trade-off between reliability and computational cost.

 Statistical tests results are provided in Supplementary, Table~\ref{tab:pvalues1}. Setup-CV results into better quality architectures than Setup-1Fold with statistical significance for SAGOMEA on the Prostate dataset with budget T and on the ACDC datasets with budgets 2T, 4T, and 8T.
 
\begin{figure}[h]
    \centering
    \includegraphics[width=0.5\textwidth]{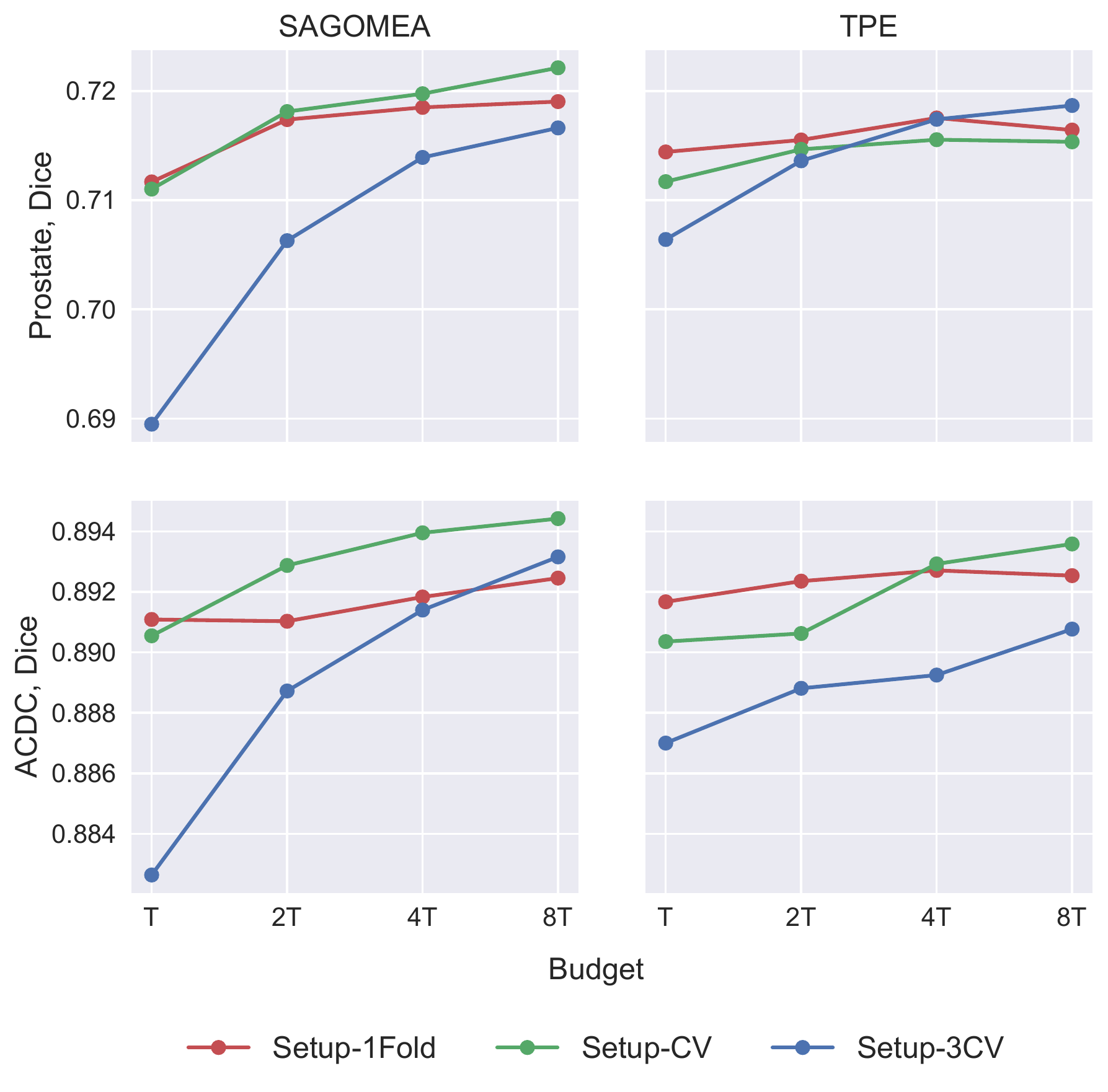}
     \caption{Main experimental results. The graphs show average (over 5 runs, 5 best architectures per run) architectures quality (obtained in the independent evaluation) under different computational budgets and different performance evaluation Setups.}
    \label{fig:search_results}
\end{figure}

\subsection{Explaining performance differences}
\begin{figure}
\includegraphics[width=0.45\textwidth]{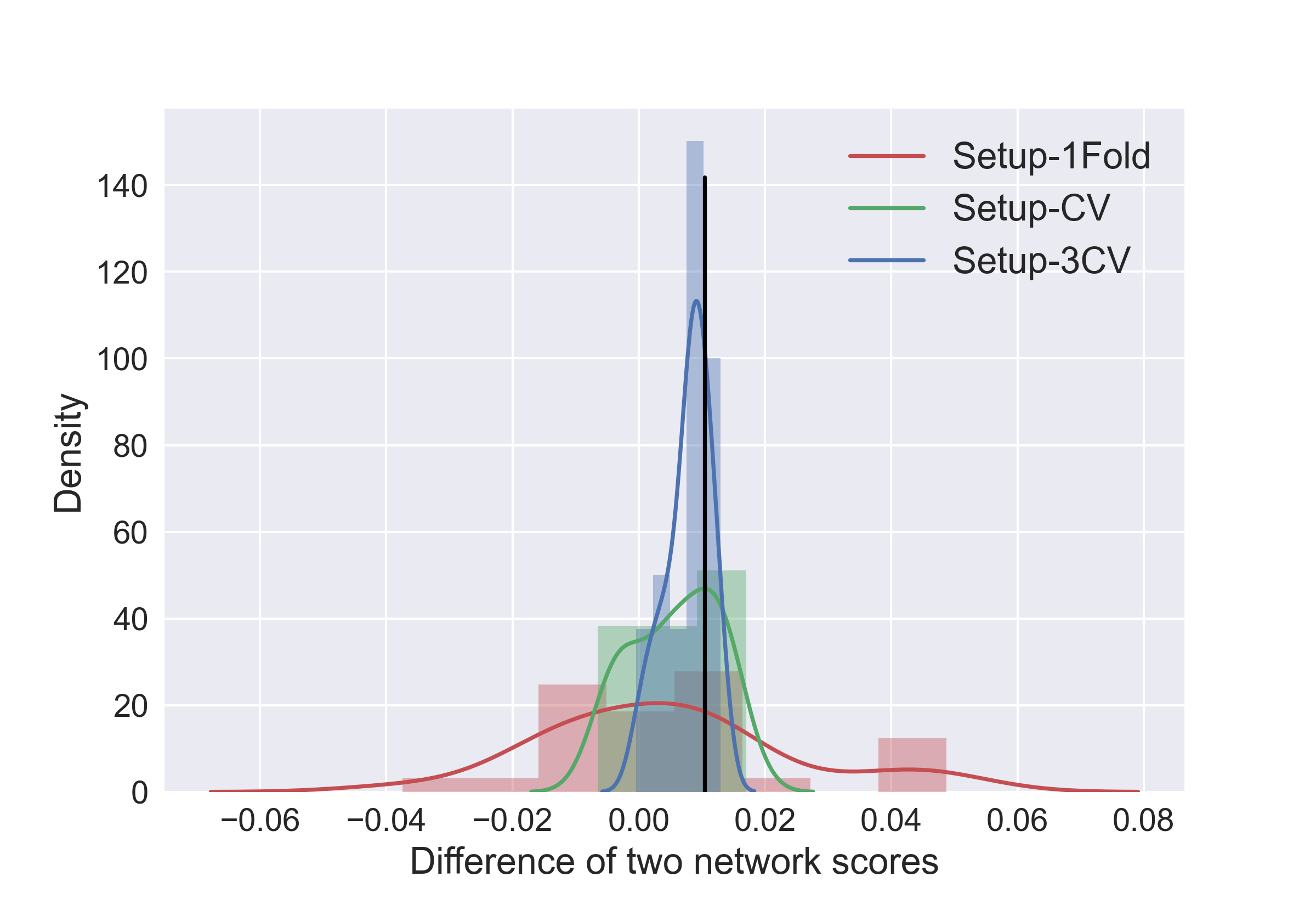}
\caption{Histograms and fitted Kernel Density Estimation (KDE) of 30 evaluation score differences (different seeds, cross-validation splits) between two networks (score of the first net minus the score of the second net) for the three investigated Setups. The black vertical line shows the performance difference in the independent evaluation procedure. The samples to the left from zero mean that these two networks are wrongly ordered: the second one is better according to the corresponding Setup, while in the independent evaluation the first one is better. The y-axis is normalized such that the area under the KDE curve is 1.}
\label{fig:distplots}
\end{figure}

In order to better understand why Setup-1Fold is underperforming, we do the following experiment. We take two networks, one is among the best found networks for the Prostate dataset, and the second also performs reasonably well, but worse than the first one by $~\approx0.01$ (in the independent evaluation). We do 30 different evaluations of these two networks (different seeds and cross-validation splits) and calculate in how many cases the first net would be preferred to the second one when different evaluation setups are used. These results are shown in Figure~\ref{fig:distplots}. When Setup-1Fold is used, the first net correctly shows better performance in only 17/30 cases (57\%). This ratio goes up to 70\% (21/30 cases) when Setup-CV is used. As expected, with Setup-3CV this ratio is even higher, namely to 29/30 cases or 97\%. While there are subtle differences in some of the scores of Setup-3CV and the independent evaluation, their average values are reasonably close. These results show that using Setup-1Fold for fitness function calculation causes many situations when a suboptimal net is selected during the search, and, therefore, such an approach underperforms compared to more reliable Setup-CV and Setup-3CV (with enough computational budget). Due to low noise in fitness function scores, Setup-3CV should potentially lead to finding the best architectures if enough computational budget is available.

\subsection{Comparison to alternative network architectures}
Though obtaining state-of-the-art networks is not the main goal of this work, we compare the performance of the found architectures to well-known Unet-like architectures in order to ensure that the used search space contains well performing architectures. The considered alternative architectures are the automatically configurable nnUnet \footnote{\href{https://github.com/MIC-DKFZ/nnUNet}{https://github.com/MIC-DKFZ/nnUNet}}, and three Unets with different encoders (Resnet-18, Efficientnet-b0, Efficientnet-b7) as implemented in the  SegmentationModelsPytorch library  \cite{Yakubovskiy:2019}  \footnote{\href{https://github.com/qubvel/segmentation\_models.pytorch}{https://github.com/qubvel/segmentation\_models.pytorch}}. 
Results are shown in Table~\ref{tab:sota}. The best found architectures are visualized in Supplementary, Figure~\ref{fig:prostatenet}. For the Prostate dataset, all our NAS configurations managed to find better networks than the alternatives. For the ACDC dataset, our Setup-CV and Setup-3CV found better networks than all alternatives except the nnUnet. The performance of nnUnet is on par with our best Setup (3CV). Note however that in contrast to our networks, nnUnet does not use downsampling in the first convolution which allows to effectively process images in higher resolution. 

\begin{table}
\caption{Comparison of the best found architectures to alternative Unet-like architectures. Numbers in the table are final (independent) evaluation scores. Note that for each of our setups, for fairness of comparison, we report here the final evaluation score of the network found to have the best fitness according to the search procedure (among all runs of all search algorithms).}
\begin{tabular}[width=\textwidth]{|l|c|c|}
 \hline
 Architecture & Prostate, Dice & ACDC, Dice   \\ \hline
 ResNet-18-Unet  & 0.690  &  0.893  \\
 EfficientNet-b0-Unet & 0.703  &  0.885  \\
 EfficientNet-b7-Unet &  0.707 & 0.895 \\ 
 nnUnet & 0.699 &   0.\textbf{897}   \\ \hline
 Ours, Setup-1Fold (best) & 0.723 & 0.894 \\  
 Ours, Setup-CV (best) & 0.723 & 0.896 \\  
 Ours, Setup-3CV (best) & \textbf{0.726} &  \textbf{0.897} \\ \hline 

\end{tabular}
\label{tab:sota}
\end{table}

\section{Discussion}
In this work we focused on NAS for medical image segmentation. Due to computational cost reasons, we used a 2D segmentation paradigm and quite compact architectures. However, it was shown \cite{isensee2018nnu}, that using a 3D segmentation approach (i.e., train on 3D volumetric patches instead of 2D patches), might be beneficial for performance. Moreover, increasing the resolution of the patches, removing the image downsampling in the stem convolution and training for more epochs might potentially substantially increase the performance of the found networks. The computational cost of our experiments and the available computing capacity did not allow us to make such modifications, but we argue that the conducted experiments are well suited for this study. 

We used a natural, yet not necessarily the most efficient type of NAS: to evaluate the performance of each architecture, we trained it from scratch. However, there exist approaches aimed at reducing the computational costs of NAS. Two main classes of such approaches are learning curve modelling (predicting architecture performance from partial training) and supernetwork-based NAS (training one large network which has all networks in the search space as its subnetworks). For practical usage of NAS, using such methods can be beneficial as they might substantially reduce the computational costs. In this work, we did not focus on such approaches. Furthermore, we observe that even without partial training (for a fewer number of epochs) network performance scores are quite noisy, and using partial training can only aggravate this problem. 
However, we believe that further studying noise in network performance evaluation with more advanced NAS techniques, including approaches based on supernetworks, is an interesting topic for further research.

\section{Conclusion}
In this work we address the problem of stochasticity in the architecture performance evaluation score in Neural Architecture Search (NAS). We focused on NAS for medical image segmentation. To reduce the stochasticity of the network performance evaluation procedure, instead of a simple performance evaluation (training a network with one random seed and evaluating it on a fixed validation set), we proposed to use cross-validation with different random seeds for each fold or even three times repeated cross-validation with different data splits. We conduct experiments on two publicly available segmentation datasets. In our experiments we allocated equal computational budget to the three considered performance evaluation setups and studied differences in performance of found networks when calculated independently from the conducted search runs. Results showed that more reliable performance evaluation setups lead to finding better performing architectures if enough computational budget is provided. We believe the obtained conclusions can be valuable for both practical applications of NAS and the development of new NAS approaches.

\begin{acks}
This work is part of the research programme Commit2Data with project number
628.011.012, which is financed by the Dutch Research Council (NWO). We thank SURF (www.surf.nl) for the support in using the Supercomputer Snellius.
\end{acks}


\clearpage
\onecolumn

\section*{Supplementary
}
\setcounter{figure}{0}   
\setcounter{table}{0}

\colorlet{significant}{blue!30}

\begin{figure*}[ht]

    \centering
    \includegraphics[height=4.2cm]{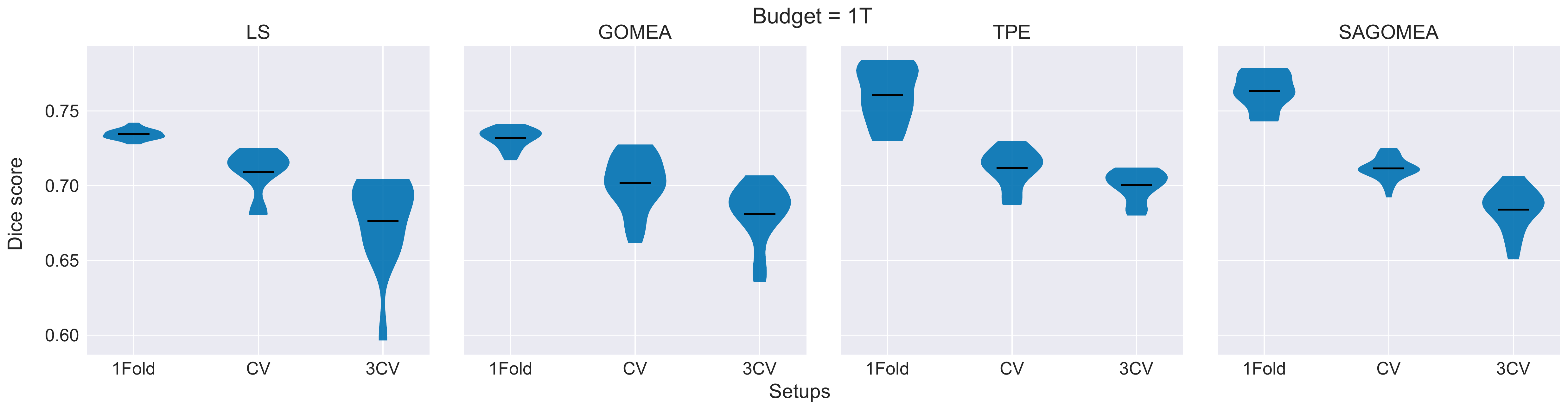} \\
    \vspace{0.5cm}
    \includegraphics[height=4.2cm]{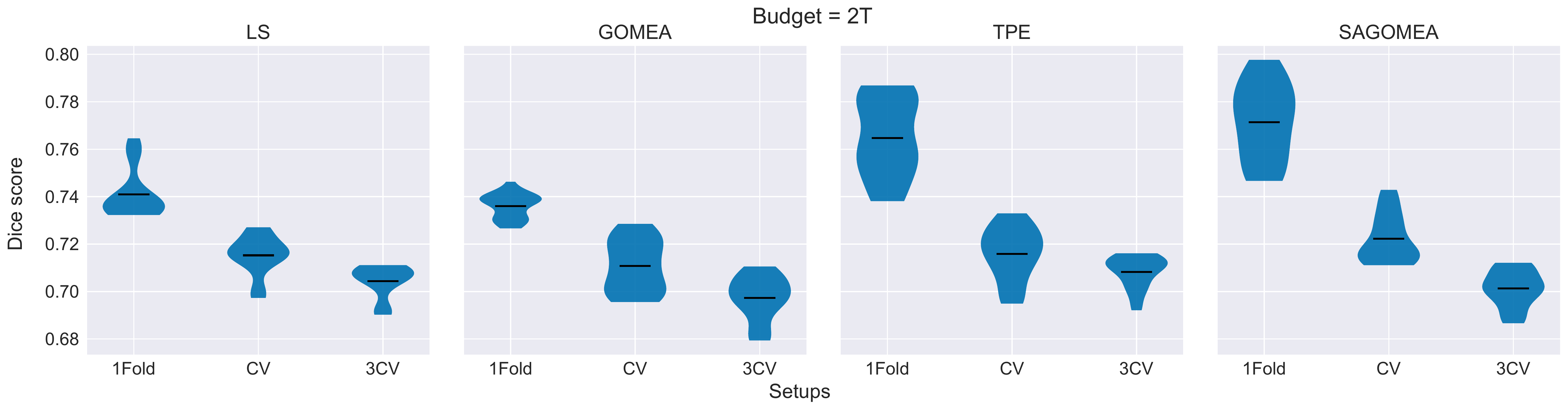} \\
    \vspace{0.5cm}
    \includegraphics[height=4.2cm]{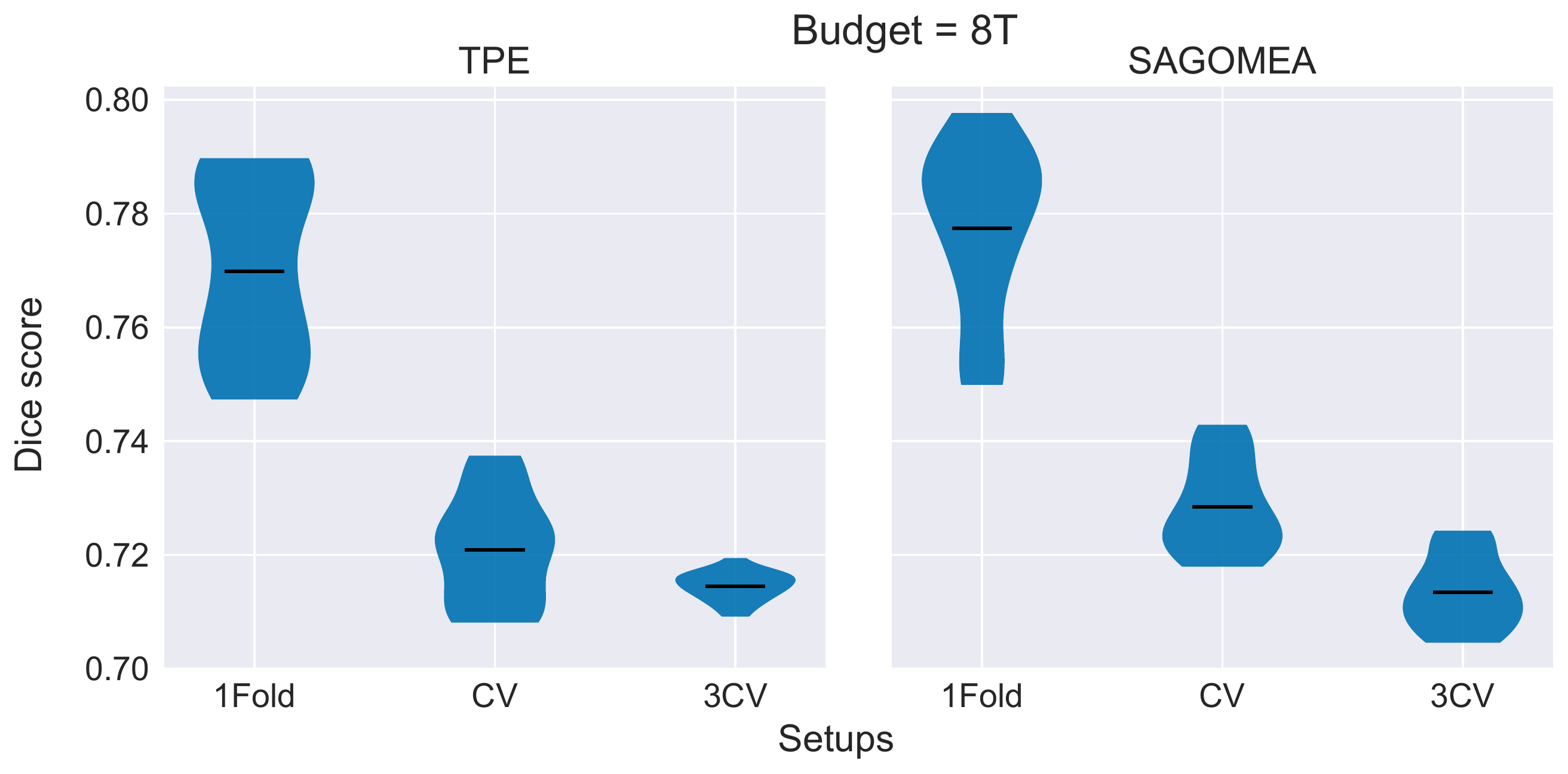}
    
    \caption{Comparing optimization performance of search algorithms on the Prostate dataset with different performance evaluation setups (i.e., fitness functions) with computational budgets T, 2T, and 8T. The image shows distributions (over five runs, five best architectures from each run) of the search results. Note that these results take into account only fitness function values, not the final architectures quality.}
    \label{fig:budgets12}

\end{figure*}

\begin{figure*}
    \centering
    \includegraphics[width=0.9\textwidth]{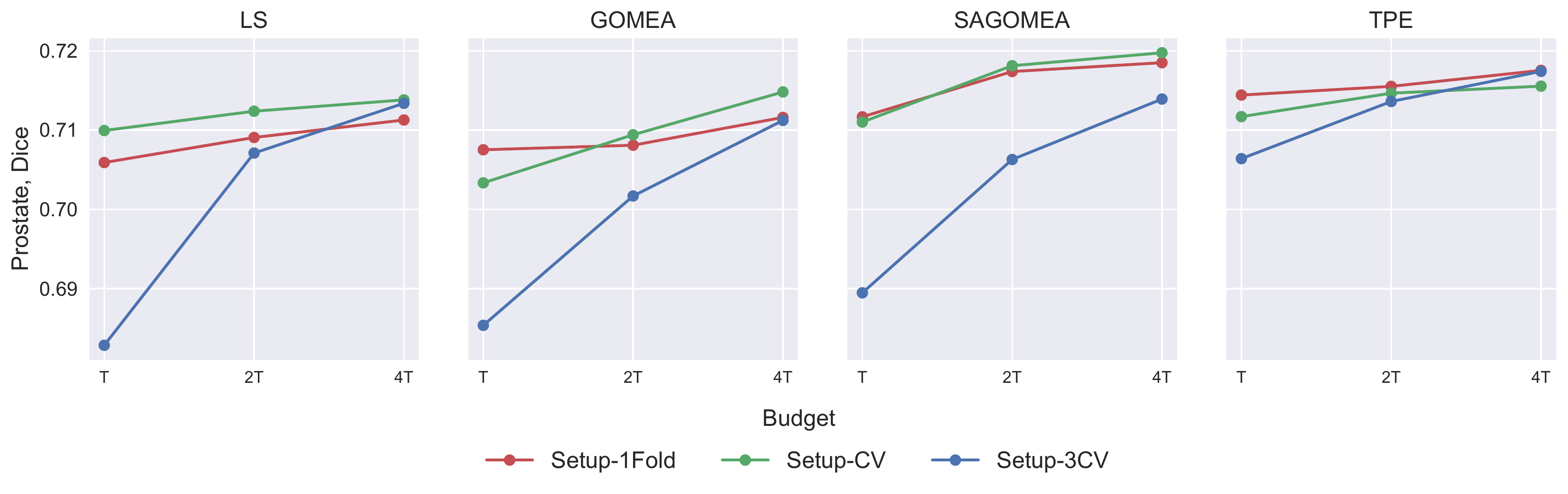}
    \caption{Experimental results for all search algorithms on the Prostate dataset. The graphs show average (over 5 runs, 5 best architectures per run) architectures quality (obtained in the independent evaluation) under different computational budgets (with up to 4T, results for budget 8T for SAGOMEA and TPE are provided in the main paper) and different performance evaluation Setups.}
    \label{fig:main_results2}
\end{figure*}
\begin{figure*}
\vspace{1cm}

    \centering
    \includegraphics[width=0.97\textwidth]{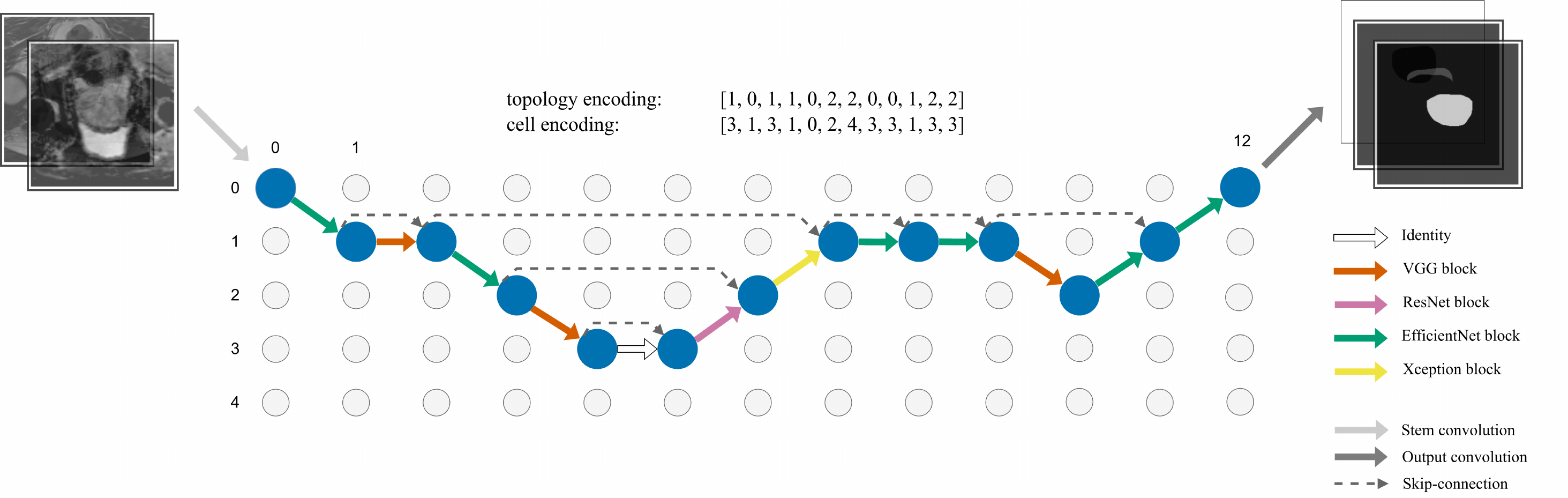} \\
    \centering
\vspace{1cm}    
    \includegraphics[width=0.97\textwidth]{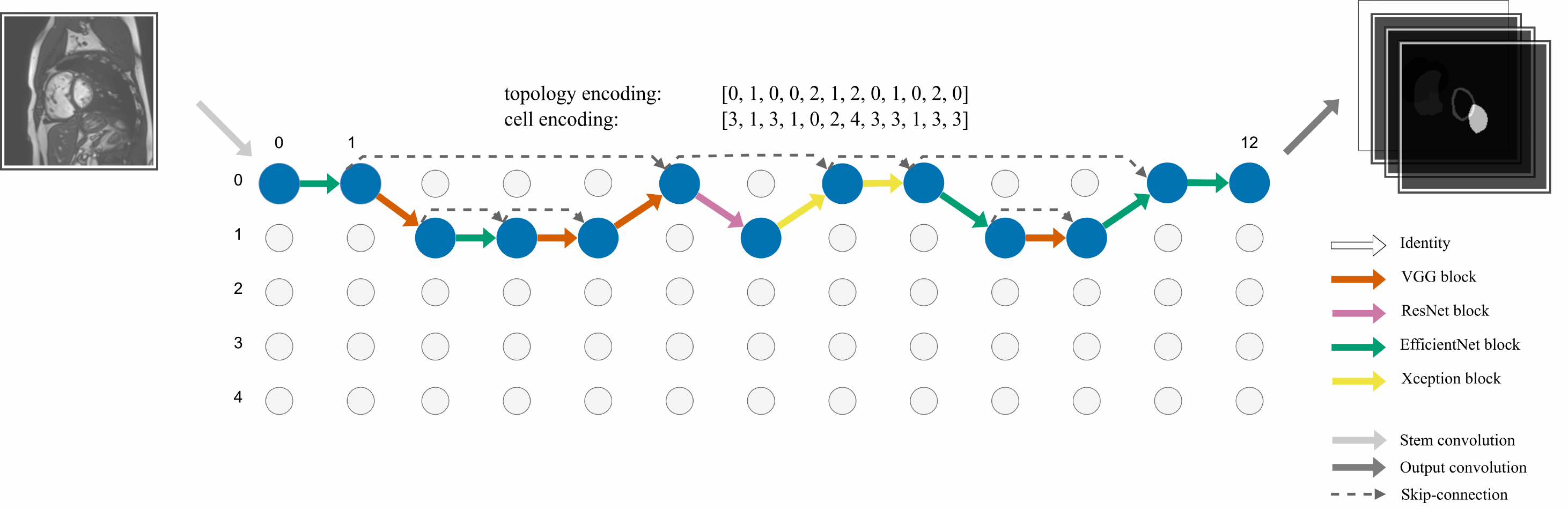}
    \caption{The best found network architecture for the Prostate (upper row) and ACDC (bottom row) datasets.}
    \label{fig:prostatenet}
\end{figure*}
 \clearpage

\begin{table*}
\centering
\caption{Experimental results for all search algorithms. The figures show average (over 5 runs, 5 best scores per run) performance evaluation scores (i.e., fitness function values) values under different computational budgets and different performance evaluation Setups. If a cell is empty, the corresponding experiment was not conducted. The best achieved performance in each row and for each setup is highlighted.}     
\label{tab:opt_results1}
\setlength{\tabcolsep}{0.3em}
\begin{tabular}{|c|c|c|c|c|c|c|c|c|c|c|c|c|c|}
\hline
& & \multicolumn{4}{c|}{\textbf{Setup-1Fold}} &  \multicolumn{4}{c|}{\textbf{Setup-CV}} &  \multicolumn{4}{c|}{\textbf{Setup-3CV}} \\ \hline
\textbf{Dataset} & \textbf{Budget} &  
\textbf{LS} &  \textbf{GOMEA} &  \textbf{TPE} &  \textbf{SAGOMEA} &
\textbf{LS} &  \textbf{GOMEA} &  \textbf{TPE} &  \textbf{SAGOMEA} &
\textbf{LS} &  \textbf{GOMEA} &  \textbf{TPE} &  \textbf{SAGOMEA} 

                    \\ \hline

\multirow{4}{*}{\textbf{Prostate}} & \textbf{T} & 0.734  & 0.732  & 0.761  & \textbf{0.763}  & 0.709  & 0.702  & \textbf{0.712}  & 0.711  & 0.676  & 0.681  & \textbf{0.700}  & 0.684 \\
& \textbf{2T} & 0.741  & 0.736  & 0.765  & \textbf{0.771}  & 0.715  & 0.711  & 0.716  & \textbf{0.722}  & 0.704  & 0.697  & \textbf{0.708}  & 0.701 \\
& \textbf{4T} & 0.744  & 0.740  & 0.767  & \textbf{0.775}  & 0.719  & 0.716  & 0.718  & \textbf{0.726}  & 0.710  & 0.707  & \textbf{0.712}  & 0.709 \\
& \textbf{8T} & 0.744  & 0.740  & 0.770  & \textbf{0.777}  & 0.719  & 0.716  & 0.721  & \textbf{0.728}  & 0.711  & 0.707  & \textbf{0.714}  & 0.713 \\ \hline
\multirow{4}{*}{\textbf{ACDC}} & \textbf{T} & -  & -  & 0.895  & \textbf{0.896}  & -  & -  & 0.891  & \textbf{0.892}  & -  & -  & \textbf{0.888}  & 0.883 \\
& \textbf{2T} & -  & -  & \textbf{0.896}  & \textbf{0.896}  & -  & -  & 0.892  & \textbf{0.895}  & -  & -  & \textbf{0.889}  & \textbf{0.889} \\
& \textbf{4T} & -  & -  & \textbf{0.897}  & \textbf{0.897}  & -  & -  & 0.894  & \textbf{0.895}  & -  & -  & 0.890  & \textbf{0.892} \\
& \textbf{8T} & -  & -  & \textbf{0.898}  & \textbf{0.898}  & -  & -  & 0.895  & \textbf{0.896}  & -  & -  & 0.892  & \textbf{0.894} \\

          \hline

\end{tabular}
\vspace{1cm}

\end{table*}

\begin{table*}
\caption{Experimental results for all search algorithms. The figures are average (over 5 runs, 5 best architectures per run) architectures quality (obtained in the independent evaluation) under different computational budgets and different performance evaluation Setups. If a cell is empty, the corresponding experiment was not conducted. The best achieved average architecture quality in each row is highlighted.}     
\label{tab:quality_results}
\setlength{\tabcolsep}{0.3em}
\begin{tabular}{|c|c|c|c|c|c|c|c|c|c|c|c|c|c|}
\hline
& & \multicolumn{4}{c|}{\textbf{Setup-1Fold}} &  \multicolumn{4}{c|}{\textbf{Setup-CV}} &  \multicolumn{4}{c|}{\textbf{Setup-3CV}} \\ \hline
\textbf{Dataset} & \textbf{Budget} &  
\textbf{LS} &  \textbf{GOMEA} &  \textbf{TPE} &  \textbf{SAGOMEA} &
\textbf{LS} &  \textbf{GOMEA} &  \textbf{TPE} &  \textbf{SAGOMEA} &
\textbf{LS} &  \textbf{GOMEA} &  \textbf{TPE} &  \textbf{SAGOMEA} 

                    \\ \hline

\multirow{4}{*}{\textbf{Prostate}} & \textbf{T} & 0.706  & 0.708  & \textbf{0.714}  & 0.712  & 0.710  & 0.703  & 0.712  & 0.711  & 0.683  & 0.685  & 0.706  & 0.689 \\
& \textbf{2T} & 0.709  & 0.708  & 0.716  & 0.717  & 0.712  & 0.709  & 0.715  & \textbf{0.718}  & 0.707  & 0.702  & 0.714  & 0.706 \\
& \textbf{4T} & 0.711  & 0.712  & 0.718  & 0.719  & 0.714  & 0.715  & 0.716  & \textbf{0.720}  & 0.713  & 0.711  & 0.717  & 0.714 \\
& \textbf{8T} & 0.711  & 0.712  & 0.716  & 0.719  & 0.714  & 0.715  & 0.715  & \textbf{0.722}  & 0.715  & 0.711  & 0.719  & 0.717 \\ \hline
\multirow{4}{*}{\textbf{ACDC}} & \textbf{T} & -  & -  & \textbf{0.892}  & 0.891  & -  & -  & 0.890  & 0.891  & -  & -  & 0.887  & 0.883 \\
& \textbf{2T} & -  & -  & 0.892  & 0.891  & -  & -  & 0.891  & \textbf{0.893}  & -  & -  & 0.889  & 0.889 \\
& \textbf{4T} & -  & -  & 0.893  & 0.892  & -  & -  & 0.893  & \textbf{0.894}  & -  & -  & 0.889  & 0.891 \\
& \textbf{8T} & -  & -  & 0.893  & 0.892  & -  & -  & \textbf{0.894}  & \textbf{0.894}  & -  & -  & 0.891  & 0.893 \\

          \hline

\end{tabular}
\vspace{1cm}
\end{table*}

\begin{table*}
\caption{P-values of Wilcoxon tests which test hypotheses that one Setup finds networks with better quality (after independent evaluation) than another one. Statistically significant results at $\alpha=0.05$ with Bonferroni correction ($m=8$, for two used datasets and four search algorithms) are highlighted. If a cell is empty, the corresponding experiment was not conducted. Values are rounded to three decimals.}     
\label{tab:pvalues1}

\begin{tabularx}{0.9\textwidth}{|c|c|c|c|c|c|c|c|c|c|}
\hline
& & \multicolumn{4}{X|}{\textbf{Setup-CV is better than Setup-1Fold}} &  \multicolumn{4}{X|}{\textbf{\textbf{Setup-3CV is better than Setup-1Fold}}} \\ \hline
      \textbf{Dataset} & \textbf{Budget}
      &  \textbf{LS} & \textbf{GOMEA} &\textbf{TPE} &   \textbf{SAGOMEA} 
      & \textbf{LS} & \textbf{GOMEA} &\textbf{TPE} &   \textbf{SAGOMEA}    \\ \hline

\multirow{4}{*}{\textbf{Prostate}} & \textbf{T} & 0.086  & 0.967  & 0.810  & 0.500  & 1.000  & 1.000  & 1.000  & 1.000 \\
& \textbf{2T} & 0.082  & 0.468  & 0.702  & 0.245  & 0.879  & 0.974  & 0.856  & 1.000 \\
& \textbf{4T} & 0.198  &  0.014  & 0.926  & 0.221  & 0.584  & 0.553  & 0.553  & 0.914 \\
& \textbf{8T} & 0.198  & \ 0.014  & 0.644  & \cellcolor{significant} 0.001  & 0.221  & 0.553  &0.026  & 0.922 \\
 \hline\multirow{4}{*}{\textbf{ACDC}} & \textbf{T} & - & - & 0.999  & 0.933  & - & - & 1.000  & 1.000 \\
& \textbf{2T} & - & - & 0.998  & \cellcolor{significant} 0.000  & - & - & 1.000  & 1.000 \\
& \textbf{4T} & - & - & 0.458  & \cellcolor{significant} 0.000  & - & - & 0.999  & 0.862 \\
& \textbf{8T} & - & - & 0.063  & \cellcolor{significant} 0.000  & - & - & 0.996  & 0.169 \\
 \hline

\end{tabularx}
    
\vspace{2cm}

\end{table*}

\begin{table*}
\caption{P-values of Wilcoxon tests which test hypotheses that one search algorithms finds better solutions (by fitness values only, not in the independent evaluation) than another one. Statistically significant results $\alpha=0.05$ with Bonferroni correction ($m=6$, for two used datasets and three Setups) are highlighted. If a cell is empty, the corresponding experiment was not conducted. Values are rounded to three decimals.}     
\label{tab:pvalues2}
\setlength\tabcolsep{1.5pt} 
\begin{tabularx}{\textwidth}{|c|c|c|c|c|c|c|c|c|c|c|c|c|c|c|c|c|}
\hline
& & \multicolumn{3}{X|}{\textbf{TPE is better  than LS}} &  \multicolumn{3}{X|}{\textbf{TPE is better than GOMEA}} &  \multicolumn{3}{X|}{\textbf{SAGOMEA is better than LS}}  &  \multicolumn{3}{X|}{\textbf{SAGOMEA is better than GOMEA}} &  \multicolumn{3}{X|}{\textbf{SAGOMEA is better than TPE}} \\ \hline
      \textbf{Dataset} & \textbf{Budget}
      &  \textbf{1Fold} & \textbf {CV} & \textbf{3CV}
      &  \textbf{1Fold} & \textbf {CV} & \textbf{3CV}
      &  \textbf{1Fold} & \textbf {CV} & \textbf{3CV}
      &  \textbf{1Fold} & \textbf {CV} & \textbf{3CV}
      &  \textbf{1Fold} & \textbf {CV} & \textbf{3CV}
      \\ \hline

\multirow{4}{*}{\textbf{Prostate}} & \textbf{T} & \cellcolor{significant} 0.000  & 0.045  & \cellcolor{significant} 0.000  & \cellcolor{significant} 0.000  & \cellcolor{significant} 0.000  & \cellcolor{significant} 0.000  & \cellcolor{significant} 0.000  & 0.895  & 0.183  & \cellcolor{significant} 0.000  & 0.070  & 0.674  & 0.067  & 0.702  & 1.000 \\
& \textbf{2T} & \cellcolor{significant} 0.000  & 0.298  & \cellcolor{significant} 0.000  & \cellcolor{significant} 0.000  & \cellcolor{significant} 0.000  & \cellcolor{significant} 0.000  & \cellcolor{significant} 0.000  & \cellcolor{significant} 0.000  & 0.862  & \cellcolor{significant} 0.000  & \cellcolor{significant} 0.000  & \cellcolor{significant} 0.003  & \cellcolor{significant} 0.000  & \cellcolor{significant} 0.000  & 0.998 \\
& \textbf{4T} & \cellcolor{significant} 0.000  & 0.720  & \cellcolor{significant} 0.004  & \cellcolor{significant} 0.000  & \cellcolor{significant} 0.000  & \cellcolor{significant} 0.000  & \cellcolor{significant} 0.000  & \cellcolor{significant} 0.000  & 0.874  & \cellcolor{significant} 0.000  & \cellcolor{significant} 0.000  & 0.205  & \cellcolor{significant} 0.000  & \cellcolor{significant} 0.000  & 0.949 \\
& \textbf{8T} & - & - & - & - & - & - & - & - & - & - & - & - & \cellcolor{significant} 0.000  & \cellcolor{significant} 0.000  & 0.831 \\
 \hline\multirow{4}{*}{\textbf{ACDC}} & \textbf{T} & - & - & - & - & - & - & - & - & - & - & - & - & 0.057  & \cellcolor{significant} 0.001  & 1.000 \\
& \textbf{2T} & - & - & - & - & - & - & - & - & - & - & - & - & 0.604  & \cellcolor{significant} 0.000  & 0.427 \\
& \textbf{4T} & - & - & - & - & - & - & - & - & - & - & - & - & 0.817  & \cellcolor{significant} 0.001  & \cellcolor{significant} 0.001 \\
& \textbf{8T} & - & - & - & - & - & - & - & - & - & - & - & - & 0.074  & \cellcolor{significant} 0.000  & \cellcolor{significant} 0.000 \\
 \hline

\end{tabularx}
    
\vspace{2cm}

\end{table*}
\clearpage
\bibliographystyle{ACM-Reference-Format}
\bibliography{bibliography}

\end{document}